\documentclass[journal]{IEEEtran}

\usepackage{amsmath, amssymb}
\usepackage{booktabs}
\usepackage{multirow}
\usepackage{graphicx}
\usepackage{overpic} %
\usepackage{enumitem}
\usepackage[dvipsnames,table]{xcolor}
\usepackage{arydshln} %
\usepackage{comment}
\usepackage{balance}
\usepackage[most]{tcolorbox}
\usepackage[numbers,sort&compress]{natbib}
\usepackage[pagebackref=false,breaklinks=true,colorlinks,bookmarks=false]{hyperref}

\newcommand{\coolname}{\textit{GemNav}}

\title{GemNav: Discrete-Token Visual {Robot} Navigation {using} a Multimodal {Large} Language Model}

\author{Peter~B\"{o}hm,
        Saimunur~Rahman,
        Abdelwahed~Khamis,
        Sagun~Man~Singh~Shrestha,
        Chris~McCool,
        and~Peyman~Moghadam%
  \thanks{The authors are with CSIRO Robotics, Brisbane, Australia.
          Corresponding author: Peter B\"{o}hm (\texttt{peter.bohm@csiro.au}).}%
}

\begin{document}

\bstctlcite{IEEEexample:BSTcontrol}

  \maketitle

  \begin{abstract}

Visual navigation policies built on large pretrained models increasingly rely on large cross-embodiment datasets and expressive action-generation mechanisms.
In this paper, we ask whether a pretrained Multimodal Large Language Model (MLLM) can be adapted efficiently for robot navigation by rethinking the representation and construction of navigation actions. We introduce \coolname{}, a visual navigation policy that represents each robot-local waypoint coordinate as a categorical distribution over discrete metric value bins and soft-decodes it into a continuous coordinate for execution. This probabilistic representation retains likelihood information over alternative coordinate values while providing a compact action space suitable for robot navigation. We further construct training trajectories using metric-distance goal sampling rather than fixed future frame offsets, exposing the policy to varying physical navigation horizons independent of dataset frame rate and robot speed. \coolname{} is trained on a single 8.7-hour public SCAND dataset, roughly three orders of magnitude smaller than competing cross-embodiment datasets. Extensive offline and closed-loop experiments show improved out-of-distribution generalization over frame-based sampling and substantially more reliable obstacle navigation than a matched continuous
regression head. The same policy transfers without per-platform retraining across unseen environments and three physically distinct robot platforms.

\end{abstract}

  \begin{IEEEkeywords}
    Vision-Based Navigation, Deep Learning for Visual Perception, Representation Learning
  \end{IEEEkeywords}

  \IEEEpeerreviewmaketitle

  \section{Introduction}

Robot navigation has traditionally been built around a tightly coupled stack of perception, state estimation, and control, tuned for a particular robot, sensor setup, or environment \cite{firoozi2025foundation}.
Pretrained Vision-Language-Action (VLA) models and visual navigation
foundation models have shifted this framing, treating navigation as a cross-embodiment learning problem: a single policy trained on hundreds to thousands of hours of trajectories across robots, environments, and sensors~\cite{shah2023vint, sridhar2024nomad, zhang2024navfom, hirose2025omnivla, doshi2025scaling}.

Recent visual navigation policies built on large pretrained models follow a common recipe: a strong visual backbone is paired with a continuous-action regression head and supervised on a union of multiple navigation datasets collected across embodiments
~\cite{hirose2025omnivla, zhang2024navfom, kim2025openvlaoft}.
Yet the scalability of this approach raises a deeper question about where the true source of generalization lies.
These policies achieve expansive transferability, but their success has come bundled with an ever-growing need for large demonstration data, specialized action heads, and visual encoders trained from scratch on robotic data. Recent benchmarking further suggests that the visual encoder is the dominant bottleneck for downstream VLA performance~\cite{zhang2026vlm4vla, dey2025revla}, pointing toward a trajectory of increasing visual scale as the primary source for improvement.

In this paper, we take a different view and ask whether pretrained multimodal large language models can be adapted efficiently for robot navigation by rethinking the representation and construction of navigation actions.
Our motivation is that multimodal large language models~\cite{gemma4_2026, zitkovich2023rt, yuanrobopoint, driess2023palm} provide strong representations that can be adapted to downstream embodied tasks.
We ground this question in short-to-medium-horizon waypoint navigation: a higher-level planner supplies the local goal as a 2D pose, a goal image, or both, and the policy predicts a short robot-frame trajectory before the planner re-queries. This setting covers most robot deployments, defines success at the scale of approximately one meter, and pairs naturally with conventional collision avoidance and re-planning without the learned policy reasoning about global topology.

To investigate this, we present \coolname{}, a visual navigation policy that adapts a pretrained Multimodal Large Language Model (MLLM) for robot navigation using low-rank adaptation across its vision and language towers. Rather than directly regressing continuous waypoint coordinates, \coolname{} represents each coordinate as a categorical distribution over discrete metric value bins and decodes these distributions back into continuous waypoints for execution. This probabilistic formulation preserves information about the relative likelihood of alternative coordinate values before a trajectory is committed for execution, providing a richer representation than a single point estimate from direct regression.

We further construct training trajectories by sampling goals according to metric displacement rather than fixed future frame offsets. This exposes the policy to a range of physical navigation horizons and reduces dependence on dataset-specific frame rates and demonstrator speeds.
\coolname{} is trained on a single open dataset of approximately 8.7 hours, roughly three orders of magnitude smaller than the cross-embodiment datasets used by OmniVLA~\cite{hirose2025omnivla}, with no manually collected or simulation data. We evaluate the same policy through extensive offline and closed-loop real-robot experiments across unseen environments and multiple robot embodiments, without per-platform retraining.
In summary, the main contributions of this paper are as follows:
\begin{itemize}[leftmargin=1.5em, itemsep=0pt, topsep=0pt]
\item We introduce a probabilistic categorical representation for metric robot navigation, where each waypoint coordinate is represented as a distribution over discrete metric value bins and soft-decoded into a continuous coordinate for execution. This formulation retains the likelihood of alternative coordinate values while providing a unified prediction space for waypoint and navigation-state decisions.
\item We introduce a metric-distance goal sampling strategy for constructing navigation supervision from existing demonstrations. In contrast to fixed frame-offset sampling, it exposes the policy to varying physical horizons independent of dataset frame rate and robot speed. %
        \item We demonstrate data-efficient adaptation of a pretrained MLLM using only an 8.7-hour training dataset. \coolname{} transfers zero-shot across unseen environments and robot embodiments and outperforms visual navigation and VLA baselines in closed-loop real-robot evaluation.
\end{itemize}

\begin{figure*}[t]
    \centering

    \includegraphics[width=0.95\linewidth]{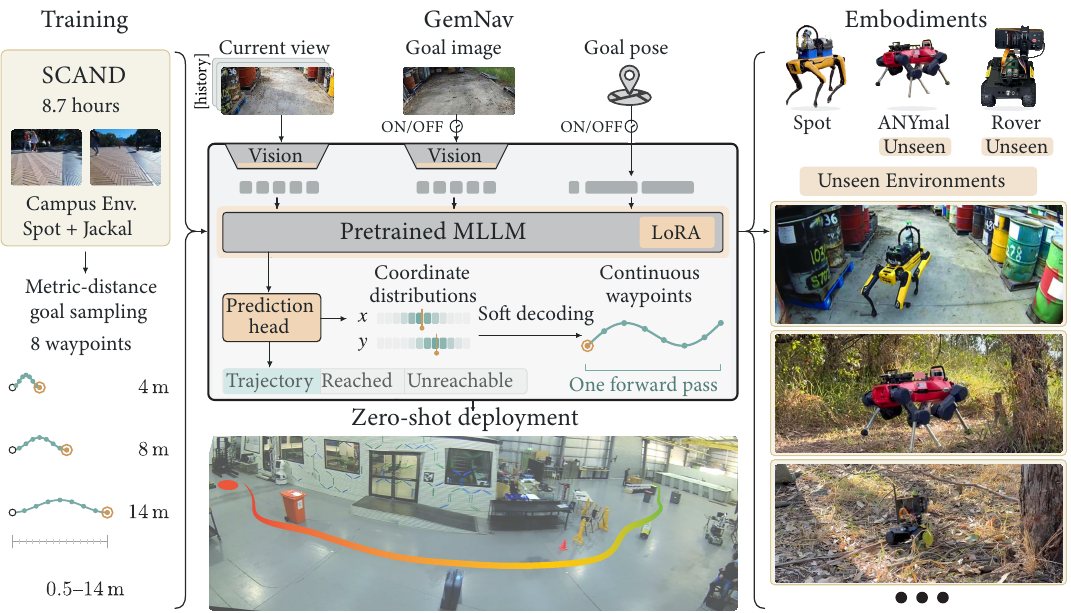}
    \caption{
\textbf{\coolname{} overview.}
\textit{Left:} Metric-distance-based goal sampling produces eight-waypoint targets over varying physical navigation horizons, reducing dependence on frame rate and demonstrator speed.
\textit{Center:} LoRA adapts the vision and language towers for image, 2D pose, or image+pose goals. A task-specific prediction head jointly produces categorical coordinate distributions for eight waypoints and a navigation state in a single forward pass; soft decoding yields continuous metric waypoints (one coordinate pair shown).
\textit{Right and bottom:} The same checkpoint transfers without per-platform retraining to unseen environments on Spot, ANYmal, and WAVE ROVER, with the latter two being unseen embodiments.
}
    \label{fig:architecture}
\end{figure*}

  \section{Related Work}
\label{sec:related-work}

\noindent\textbf{Cross-embodiment Visual Navigation:}
Navigation foundation models train dedicated policies on robot trajectories: GNM learns a shared action space across six robots~\cite{shah2023gnm}, ViNT uses a CNN--Transformer architecture~\cite{shah2023vint}, NoMaD unifies exploration and goal reaching through diffusion~\cite{sridhar2024nomad}, and NavFoM scales to 8M samples across sensor configurations~\cite{zhang2024navfom}. OmniVLA supports language, pose, and image goals via
a continuous regression head trained on 9,500 hours~\cite{hirose2025omnivla}. JanusVLN \cite{zeng2025janusvln} couples a frozen Qwen2.5-VL backbone with a frozen 3D-geometry encoder for continuous VLN.  In contrast, \coolname{} adapts a pretrained MLLM directly for visual navigation using a compact body-frame output interface, and is trained on a single 8.7-hour corpus while transferring across unseen environments and cross-embodiments.

\noindent\textbf{Navigation Supervision:} %
Existing navigation policies commonly construct training targets by selecting goals and future waypoints at fixed frame offsets along recorded demonstrations~\cite{shah2023gnm, hirose2025omnivla}. This couples the physical navigation horizon to the dataset frame rate and robot speed. Other approaches broaden navigation supervision through additional data sources: LeLaN augments action-free video with language and counterfactual actions~\cite{hirose2024lelan}, CityWalker learns urban navigation from web-scale videos~\cite{liu2025citywalker}, and CAST uses counterfactual hindsight relabeling to improve instruction following~\cite{glossop2025cast}.
\coolname{} instead focuses on how supervision is constructed from existing demonstrations, sampling goals according to metric displacement rather than a fixed number of future frames. This exposes the policy to varying physical horizons while retaining the demonstrated intermediate path.

\noindent\textbf{Action Representations for VLAs:} Recent VLAs use increasingly expressive action representations: diffusion~\cite{chi2023diffusion, hou2025dita} and flow-matching methods~\cite{black2024pi0, bjorck2025gr00t} model continuous action distributions, while FAST~\cite{pertsch2025fast} compresses action sequences into discrete tokens for efficient autoregressive training. $\pi_0$~\cite{black2024pi0} uses flow matching with a dedicated action expert, and $\pi_{0.5}$~\cite{intelligence2025pi05} combines discrete FAST tokens during pretraining with continuous flow-matching actions at inference.
These suit high-dimensional manipulation, whereas navigation needs only
low-dimensional waypoints. RT-1~\cite{brohan2022rt}, RT-2~\cite{zitkovich2023rt}
and OpenVLA~\cite{kim2025openvla} discretize each action dimension into bins
decoded as single tokens, as does UniWM~\cite{dong2026towards} for navigation,
restricting actions to fixed bin values. \coolname{}
instead adopts a distributional waypoint readout: metric bins act as support points,
and each coordinate is the probability-weighted expectation over them.
This retains the more robust training of classification
losses without quantizing the output. %

  \section{Method}
\label{sec:method}

\coolname{} formulates visual robot navigation using a probabilistic categorical representation of compact body-frame waypoint primitives. Each waypoint coordinate is represented as a distribution over discrete metric value bins, while training trajectories are constructed by sampling goals according to metric displacement rather than fixed frame offsets. The resulting policy is instantiated using a pretrained MLLM, Gemma 4~\cite{gemma4_2026}, adapted using Low-Rank Adaptation (LoRA). A task-specific prediction head jointly produces the waypoint and
navigation-state logits in a single forward pass. The predicted waypoint distributions are soft-decoded into continuous coordinates for execution, and the policy replans from the current visual context and goal (Fig.~\ref{fig:architecture}).

\subsection{Problem Formulation}
\label{subsec:formulation}

At control step $t$, the policy receives a visual context

\begin{equation}
\mathcal{I}_t^{(H)}
=
\left\{
I_{t-H},\ldots,I_{t-1},I_t
\right\},
\end{equation}
where $I_t$ is the current egocentric RGB image, $H$ is the number of optional history frames. \coolname{} supports three distinct goal-conditioning modes. In the \emph{image} goal, the goal is specified only by a goal image $I^g$; in the \emph{2D pose} goal, it is specified only by a 2D goal position $\mathbf{p}_t^g=(x_t^g,y_t^g)\in\mathbb{R}^2$; and in the \emph{image+2D pose} goal, both $I^g$ and $\mathbf{p}_t^g$ are provided. Accordingly, the goal input $g_t$ takes one of these three forms depending on the selected modality.
The goal position is represented in a local 2D Cartesian frame centered on the robot, with the $+x$ axis pointing forward and the $+y$ axis pointing to the left. At each control step, platform odometry is used only to transform the goal into the current robot frame.
Given the visual context $\mathcal{I}_t^{(H)}$ and goal input $g_t$, the policy predicts a sequence of $N$ body-frame waypoint primitives

\begin{equation}
\boldsymbol{\tau}_t
=
\left\{\mathbf{w}_{t,n}\right\}_{n=1}^{N},
\qquad
\mathbf{w}_{t,n}
=
(x_{t,n},y_{t,n})\in\mathbb{R}^2,
\end{equation}
together with a navigation-state variable
\begin{equation}
c_t \in \mathcal{C}
=
\{\texttt{traj},\texttt{goal\_reached},\texttt{goal\_unreachable}\},
\end{equation}
where $\texttt{goal\_reached}$ denotes that the robot is already at the
goal, while $\texttt{goal\_unreachable}$ denotes a visually disconnected
goal and acts as a refuse-to-plan signal.
The policy can therefore be written as
\begin{equation}
\mathcal{F}_{\theta}
\left(
\mathcal{I}_t^{(H)},g_t
\right)
\rightarrow
\left(
\hat{\boldsymbol{\tau}}_t,\hat{c}_t
\right).
\end{equation}
When $\hat{c}_t=\texttt{traj}$, the predicted waypoints are passed to the platform-specific low-level controller for execution; otherwise, the corresponding navigation state is returned.
The waypoints carry no explicit yaw, as the bearing from each waypoint to the next sets the heading and the controller follows the arc through the sequence.

\subsection{Action Representation}
\label{sec:action_rep}

Rather than directly regressing continuous waypoint coordinates, \coolname{} represents each coordinate as a categorical distribution over discrete metric
values. For a coordinate $v\in[-B,B]$, we define a uniform quantizer with $K$ bins as:

\begin{equation}
Q(v)
=
\min\left(
K-1,
\left\lfloor
\frac{v+B}{2B}K
\right\rfloor
\right),
\label{eq:quant}
\end{equation}
where the metric center of bin $j\in\{0,\ldots,K-1\}$ is
\begin{equation}
\mu_j
=
-B+
\left(j+\frac{1}{2}\right)\frac{2B}{K}.
\end{equation}

For each waypoint $\mathbf{w}_{t,n}=(x_{t,n},y_{t,n})$, the policy predicts categorical logits
$\boldsymbol{\ell}_{t,n}^{q}\in\mathbb{R}^{K}$ for each
$q\in\{x,y\}$. The corresponding bin probabilities are

\begin{equation}
\pi_{t,n,j}^{q}
=
\frac{
\exp\!\left(\ell_{t,n,j}^{q}\right)
}{
\sum_{j'=0}^{K-1}
\exp\!\left(\ell_{t,n,j'}^{q}\right)
},
\qquad
q\in\{x,y\},\,
\label{eq:bin_prob}
\end{equation}
where $\pi_{t,n,j}^{q}$ denotes the probability assigned to bin $j$ for coordinate $q$ of waypoint $n$ and $j\in\{0,\ldots,K-1\}$. The corresponding categorical training targets are $Q(x_{t,n})$ and $Q(y_{t,n})$.  Thus, each coordinate is represented by a probability distribution over discrete robot-local metric values rather than a single continuous point estimate, retaining the relative likelihood assigned to alternative values.
The categorical representation is mapped back to metric space through its
expected value,
\begin{equation}
\widetilde{w}_{t,n}^{q}
=
\sum_{j=0}^{K-1}
\mu_j\pi_{t,n,j}^{q},
\qquad q\in\{x,y\}.
\label{eq:soft_decode}
\end{equation}
This soft decoding uses the full predicted distribution to recover a
continuous coordinate for metric supervision and execution.
For pose-conditioned navigation, the goal coordinates $x_t^g$ and $y_t^g$ are quantized using the same mapping $Q(\cdot)$. Consequently, pose-goal conditioning and waypoint prediction are grounded in the same robot-local
metric discretization. The predicted waypoint distributions are converted to continuous coordinates
at inference, as described in Sec.~\ref{sec:inference}.

\subsection{Metric-Distance Trajectory Supervision}
\label{sec:trajectory_supervision}

Training supervision is constructed from teleoperated demonstrations. Let a demonstration be a sequence
\begin{equation}
\mathcal{D}
=
\left\{
(I_i,\mathbf{p}_i,\psi_i)
\right\}_{i=1}^{T},
\end{equation}
where $\mathbf{p}_i\in\mathbb{R}^2$ and $\psi_i$ denote the robot position
and heading in a common world frame at frame $i$. For a current frame $t$,
we sample a target metric displacement $\rho \sim
\mathcal{U}(\rho_{\min},\rho_{\max})$ and select the goal frame as the first future observation whose displacement
from the current position reaches $\rho$:
\begin{equation}
t_g
=
\min
\left\{
i>t
\,\middle|\,
\left\|
\mathbf{p}_i-\mathbf{p}_t
\right\|_2
\geq \rho
\right\}.
\label{eq:goal_sampling}
\end{equation}

The corresponding goal image is $I^g=I_{t_g}$. To construct the waypoint
targets, we sample $N$ frames uniformly between $t$ and $t_g$ as
\begin{equation}
t_n
=
t+
\left\lfloor
\frac{n}{N}(t_g-t)
\right\rfloor,
\qquad
n=1,\ldots,N,
\label{eq:waypoint_indices}
\end{equation}
and transform their positions into the robot frame at $t$:
\begin{equation}
\mathbf{w}_{t,n}
=
\mathbf{R}(-\psi_t)
\left(
\mathbf{p}_{t_n}-\mathbf{p}_t
\right),
\qquad
\mathbf{w}_{t,N}=\mathbf{p}_t^g,
\label{eq:waypoint_transform}
\end{equation}
where $\mathbf{R}(\cdot)\in\mathbb{R}^{2\times2}$ denotes the planar rotation matrix. Unlike fixed frame-offset sampling, this construction defines the training horizon in metric space rather than by dataset frame rate or robot speed, exposing the policy to varying physical goal distances while preserving the demonstrated trajectory. Coordinates outside $[-B,B]$ are projected onto the boundary before applying Eq.~\eqref{eq:quant}. The \texttt{goal\_reached} class is generated from near-goal samples, whereas \texttt{goal\_unreachable} uses cross-trajectory goal negatives.%

\subsection{MLLM Adaptation and Prediction Head}
\label{sec:model}

The visual context $\mathcal{I}_t^{(H)}$ and goal input $g_t$ are
assembled into a multimodal prompt $\mathcal{P}_t$ containing a fixed task
instruction, the visual observations, the goal image when present, and the
goal position when required by the selected modality. For pose-conditioned
inputs, the coordinates are represented by
\begin{equation}
\mathcal{T}_t^g
=
\left[
\texttt{<goal\_coord>},
\nu_{Q(x_t^g)},
\nu_{Q(y_t^g)}
\right],
\label{eq:goal_tokens}
\end{equation}
where $\nu_j$ denotes the reserved vocabulary token associated with metric
bin $j$. The prompt terminates with a \texttt{Trajectory:} marker from
which the prediction is obtained.

Let $f_{\theta_0,\theta_L}$ denote the pretrained MLLM with fixed backbone
parameters $\theta_0$ and trainable Low-Rank Adaptation (LoRA) parameters
$\theta_L$ applied to both the vision and language towers. The final-layer
representation at the \texttt{Traj:} marker is
\begin{equation}
\mathbf{z}_t
=
f_{\theta_0,\theta_L}(\mathcal{P}_t)_{\mathrm{traj}}
\in\mathbb{R}^{D}.
\label{eq:mllm_rep}
\end{equation}

A task-specific prediction head first transforms $\mathbf{z}_t$ as
\begin{equation}
\mathbf{u}_t
=
\operatorname{GELU}
\left(
\mathbf{W}_h\mathbf{z}_t+\mathbf{b}_h
\right),
\label{eq:pred_head}
\end{equation}
and jointly produces the waypoint and navigation-state logits,
\begin{equation}
\boldsymbol{\ell}_{t,n}^{q}
=
\mathbf{W}_{n,q}\mathbf{u}_t+\mathbf{b}_{n,q}
\in\mathbb{R}^{K},
\qquad
q\in\{x,y\},
\end{equation}
\begin{equation}
\boldsymbol{\ell}_t^{c}
=
\mathbf{W}_{c}\mathbf{u}_t+\mathbf{b}_{c}
\in\mathbb{R}^{|\mathcal{C}|}.
\label{eq:state_logits}
\end{equation}
The waypoint logits define the categorical probabilities in Eq.~\eqref{eq:bin_prob}, while $\boldsymbol{\ell}_t^{c}$ defines the navigation states in $\mathcal{C}$, with both predicted jointly in a single forward pass.

\subsection{Training Objective}
\label{sec:objective}

Let $w_{t,n}^{q}$ denote coordinate $q\in\{x,y\}$ of waypoint
$\mathbf{w}_{t,n}$, such that $w_{t,n}^{x}=x_{t,n}$ and
$w_{t,n}^{y}=y_{t,n}$. For trajectory samples, the categorical
cross-entropy over the quantized coordinates is
\begin{equation}
\mathcal{L}_{\mathrm{CE}}^{w}
=
-\frac{1}{2N}
\sum_{n=1}^{N}
\sum_{q\in\{x,y\}}
\log
\pi_{t,n,Q(w_{t,n}^{q})}^{q}.
\label{eq:traj_ce}
\end{equation}

Categorical cross-entropy does not encode the metric ordering of the bins,
since predictions in adjacent and distant bins receive the same penalty when
incorrect. Using the soft-decoded coordinate in Eq.~\eqref{eq:soft_decode}, we therefore impose an auxiliary mean squared error,
\begin{equation}
\mathcal{L}_{\mathrm{MSE}}^{w}
=
\frac{1}{2N}
\sum_{n=1}^{N}
\sum_{q\in\{x,y\}}
\left(
\widetilde{w}_{t,n}^{q}-w_{t,n}^{q}
\right)^2.
\label{eq:traj_mse}
\end{equation}

For the navigation state $c_t$, let
$\boldsymbol{\ell}_{t}^{c}\in\mathbb{R}^{|\mathcal{C}|}$ denote the
corresponding categorical logits. Its cross-entropy loss is
\begin{equation}
\mathcal{L}_{\mathrm{CE}}^{c}
=
-\ell_{t,c_t}^{c}
+
\log
\sum_{m=1}^{|\mathcal{C}|}
\exp\!\left(\ell_{t,m}^{c}\right).
\label{eq:state_ce}
\end{equation}

For trajectory samples, the full training objective is
\begin{equation}
\mathcal{L}
=
\mathcal{L}_{\mathrm{CE}}^{w}
+
\lambda_{\mathrm{MSE}}
\mathcal{L}_{\mathrm{MSE}}^{w}
+
\lambda_{c}
\mathcal{L}_{\mathrm{CE}}^{c}.
\label{eq:objective}
\end{equation}

For \texttt{goal\_reached} and \texttt{goal\_unreachable} samples,
waypoint supervision is not applied, and the objective reduces to
$\mathcal{L}=\lambda_c\mathcal{L}_{\mathrm{CE}}^{c}$. The LoRA parameters and prediction-head parameters are optimized jointly, while the pretrained backbone parameters remain fixed.

\subsection{Inference and Closed-Loop Execution}
\label{sec:inference}

At each control step, \coolname{} jointly predicts the navigation-state
logits and waypoint logits in a single forward pass. The navigation state
is selected as
\begin{equation}
\hat{c}_t
=
\arg\max_{s\in\mathcal{C}}
\ell_{t,s}^{c}.
\label{eq:nav_inference}
\end{equation}
If $\hat{c}_t=\texttt{traj}$, the waypoint logits are converted to
categorical probabilities using Eq.~\eqref{eq:bin_prob}, and each coordinate
is recovered as the expected metric value in Eq.~\eqref{eq:soft_decode}:
\begin{equation}
\hat{\mathbf{w}}_{t,n}
=
\left(
\widetilde{w}_{t,n}^{x},
\widetilde{w}_{t,n}^{y}
\right),
\qquad n=1,\ldots,N.
\end{equation}
The resulting continuous waypoint sequence is
$\hat{\boldsymbol{\tau}}_t=
(\hat{\mathbf{w}}_{t,n})_{n=1}^{N}$.
Soft decoding uses the full predicted distribution and provides continuous
metric coordinates without restricting execution to the discrete bin
centers.

The predicted waypoint sequence is passed to the platform-specific
controller for execution. If
$\hat{c}_t=\texttt{goal\_reached}$ or
$\hat{c}_t=\texttt{goal\_unreachable}$, execution is halted and the
corresponding navigation state is returned. At the next control step, the
goal is transformed into the updated robot frame and the policy replans from
the current frame. No persistent map or recurrent state is carried
between control steps.

\begin{table}[t]
\centering
\footnotesize
\setlength{\tabcolsep}{2pt}
\caption{\coolname{} main hyperparameters. }
\label{tab:hyperparams}
\begin{tabular}{@{}llll@{}}
\toprule
\multicolumn{2}{@{}l}{\textbf{Action representation}} & \multicolumn{2}{l@{}}{\textbf{Training}} \\
\midrule
Waypoints $N$ & $8$ & $\lambda_{\mathrm{MSE}}$ & $0.1$ \\
Bins per coordinate $K$ & $64$ & $\lambda_{c}$ & $0.1$ \\
Coordinate range $B$ & $15$\,m & Optimizer & AdamW \\
Goal distance $[\rho_{\min},\rho_{\max}]$ & $[0.5,14]$\,m & Learning rate & $10^{-4}$ \\
\cmidrule(r){1-2}
\multicolumn{2}{@{}l}{\textbf{Model and adaptation}} & Training steps & $40{,}000$/stage \\
\cmidrule(r){1-2}
Backbone & Gemma-4-E2B-it & & \\
LoRA rank $r$ & $32$ & & \\
LoRA scope & vision + language & & \\
Prediction-head width & $1024$ & & \\
\bottomrule
\end{tabular}
\end{table}

  \section{Experimental Setup}
\label{sec:experiments}

\subsection{Training and Evaluation Data}

We train \coolname{} only on SCAND~\cite{karnan2022socially}, an 8.7-hour public dataset of teleoperated navigation collected with Boston Dynamics Spot and Clearpath Jackal robots. We use only RGB images and robot poses; LiDAR, operator commands, and social-navigation annotations are not used. Of the 120 recordings, 108 are used for training and 12 are held out entirely for validation, yielding approximately 243k training and 29k validation examples.

Offline evaluation uses three datasets with increasing distribution shift. For in-distribution evaluation, we report results on a fixed 500-sample subset of the SCAND validation set, balanced across the three goal
modalities.
TokenWalker is a held-out dataset we collected specifically using Spot with a forward-facing Intel RealSense camera. It comprises five sessions recorded over three days, totaling 83 minutes and 9,924 frames across paved roads, sidewalks, and an industrial yard. It differs from SCAND in camera, frame rate, and environment, providing a deployment-aligned OOD test. We evaluate a fixed 500-sample modality-balanced subset.
We further evaluate on the public WildCross~\cite{wildcross2026} dataset, a held-out natural environment dataset containing eight traversals across two forest environments over 33\,km. Its repeated, reverse, and extended routes span substantial viewpoint and appearance changes in dense vegetation and irregular terrain. We use only the RGB images and 6DoF poses, reduced to $(x,y,\psi)$. After removing static segments and subsampling to 10\,Hz, approximately 271k frames remain; evaluation uses a fixed 1,200-example subset. WildCross therefore provides a stronger natural-domain and sensing domain shift than TokenWalker.

\subsection{Zero-Shot Real-Robot Deployment}

We evaluate the same SCAND-trained checkpoint in closed loop on three physically distinct robots: Spot, ANYmal D, and a WaveShare WAVE ROVER. ANYmal and the rover are unseen embodiments, while the deployment Spot uses a different camera and sensing configuration from the Spot used in SCAND. No policy parameters are adapted to any robot or deployment environment. Each platform provides its own body-frame pose estimation and low-level waypoint controller, while the \coolname{} weights and inference interface remain unchanged. All closed-loop inference is run on-board on an NVIDIA Jetson AGX Thor (128\,GB). Table~\ref{tab:platforms} summarizes the platform variation.

The zero-shot, real-robot deployment spans five settings with increasing navigation and domain shift challenges. The \emph{Open Approach} is a clear 10--12\,m paved outdoor route used to test goal reaching from different initial headings. The \emph{Detour Course} extends this to approximately 18\,m and places obstacles across the direct route, requiring the robot to deviate from the goal direction and subsequently recover it. The \emph{Warehouse} uses approximately 32\,m indoor routes with the goal outside the initial field of view and evaluates both clear and increasingly obstructed paths. The outdoor \emph{Chemical Yard} spans approximately 30--34\,m through narrow passages and dead ends, with no direct route to the goal. Finally, an \emph{off-road} deployment evaluates Spot and ANYmal on 25--30\,m natural, hillside routes through open woodland, with trees as obstacles and approximately 1.7\,m elevation gain.

The \emph{Open Approach} evaluates image, 2D pose, and image+2D pose goals, while the more challenging courses primarily use pose-conditioned goals; the \emph{off-road} deployment uses an image+2D pose goal. For image-only trials, the goal image is captured once from the robot camera at the target pose and reused across the corresponding trials. Goal coordinates are expressed in the robot frame and updated using platform odometry during execution. Model--modality--setting combinations are repeated over five trials unless stated otherwise.

\begin{table}[t]
\centering
\caption{Training and deployment platforms. Height $h$ is above ground. $v$ is platform average speed in $(m/s)$}
\label{tab:platforms}
\scriptsize
\setlength{\tabcolsep}{2.5pt}
\begin{tabular}{@{}lllll@{}}
\toprule
\textbf{Platform} & \textbf{Role} & \textbf{Camera $(h$/FoV)} & \textbf{Pose} & \textbf{$v$ $(m/s)$} \\
\midrule
Spot
& Train
& Kinect ($\sim$1.45\,m/90$^\circ$)
& Kin.+vis.
& 1.38 \\

Jackal
& Train
& Kinect ($\sim$0.47\,m/90$^\circ$)
& Wheel odom.
& -- \\
\addlinespace[2pt]
\hdashline
\addlinespace[2pt]
Spot
& Test
& D435 (1.15\,m/69$^\circ$)
& SLAM
& 0.62 \\

ANYmal D
& Test
& D455 (1.0\,m/90$^\circ$)
& Onboard
& 0.65 \\

WAVE ROVER
& Test
& Pi Cam 3 (0.115\,m/66$^\circ$)
& Phone VIO
& 0.41 \\
\bottomrule
\end{tabular}
\end{table}

\subsection{Baselines}

We compare \coolname{} with GNM~\cite{shah2023gnm}, ViNT~\cite{shah2023vint}, NoMaD~\cite{sridhar2024nomad}, and
OmniVLA~\cite{hirose2025omnivla}. GNM, ViNT, and NoMaD are evaluated in their image-goal setting using their released checkpoints and the same camera input and waypoint controller as \coolname{}. OmniVLA is a
cross-embodiment VLA supporting image and pose-conditioned goals and is evaluated using its released checkpoint and inference interface. SCAND is included in the training data of GNM, ViNT, and NoMaD, whereas
\coolname{} is trained only on SCAND.
Neither \coolname{} nor any baseline is fine-tuned on any evaluation environment; all models are evaluated with fixed weights.

\subsection{Metrics}

Offline trajectory prediction is evaluated using Average Displacement Error (ADE) and Final Displacement Error (FDE) over the predicted waypoints.
For closed-loop evaluation, each model and setting is run for five trials unless stated otherwise. A \coolname{} trial ends when it reports \texttt{goal\_reached}. We report the number of trials that reach the goal and the final goal displacement $\Delta$ at the stop. For baselines without a stopping mechanism, $\Delta$ is measured at the closest approach to the goal.

  \section{Results}
\label{sec:result}

\begin{table*}[h!]
  \centering
  \footnotesize
  \setlength{\tabcolsep}{5pt}
  \renewcommand{\arraystretch}{0.9}
  \caption{Goal displacement $\Delta$ (m, mean$\pm$std) across goal modalities, five trials per method and setting. Dashes mark settings not run.} %

  \label{tab:modality}
  \begin{tabular}{ll cc cc cc}
    \toprule
    & & \multicolumn{2}{c}{Image} & \multicolumn{2}{c}{2D Pose} & \multicolumn{2}{c}{Image + 2D Pose} \\
    \cmidrule(lr){3-4} \cmidrule(lr){5-6} \cmidrule(lr){7-8}
    Difficulty & Setting & OmniVLA & \coolname{} & OmniVLA & \coolname{} & OmniVLA & \coolname{} \\
    \midrule
    \multirow{2}{*}{Easy} & \emph{Open Approach} (A) & $1.77{\pm}0.07$ & $\mathbf{0.41{\pm}0.05}$ & $\mathbf{0.32{\pm}0.07}$ & $0.43{\pm}0.04$ & $\mathbf{0.22{\pm}0.09}$ & $0.42{\pm}0.06$ \\
    & \emph{Open Approach} (B) & $1.13{\pm}0.28$ & $\mathbf{0.20{\pm}0.05}$ & $0.48{\pm}0.06$ & $\mathbf{0.39{\pm}0.02}$ & $0.91{\pm}0.22$ & $\mathbf{0.42{\pm}0.04}$ \\
    \midrule
    \multirow{2}{*}{Medium} & \emph{Detour Course} ($90^\circ$) & -- & -- & $3.23{\pm}1.98$ & $\mathbf{0.42{\pm}0.02}$ & $5.13{\pm}0.10$ & $\mathbf{0.40{\pm}0.03}$ \\
    & \emph{Warehouse} (clear) & -- & -- & -- & -- & $0.51{\pm}0.07$ & $\mathbf{0.46{\pm}0.03}$ \\
    \midrule
    \multirow{2}{*}{Hard} & \emph{Warehouse} (blocked) & -- & -- & -- & -- & $15.46{\pm}0.12$ & $\mathbf{0.44{\pm}0.03}$ \\
    & \emph{Chemical Yard} & -- & -- & $15.89{\pm}0.30$ & $\mathbf{0.44{\pm}0.03}$ & $16.54{\pm}0.02$ & $\mathbf{0.49{\pm}0.02}$ \\
    \bottomrule
  \end{tabular}
  \vspace{-0.5em}
\end{table*}

\subsection{Zero-Shot Image-Goal Navigation}
\label{sec:image_goal_results}

\begin{figure}[t]
  \centering
  \includegraphics[width=\linewidth]{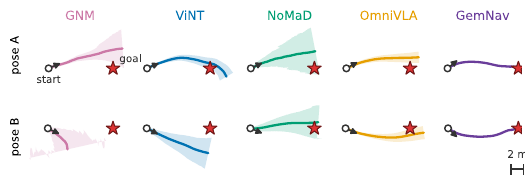}
  \caption{Image-goal navigation in the \emph{Open Approach}
on Spot. Mean trajectories and variation over five trials
per policy and initial heading.}
  \label{fig:baselines}
\end{figure}

We first evaluate image-goal navigation in the obstacle-free \emph{Open Approach}, providing a common comparison with GNM, ViNT, NoMaD, and OmniVLA, as these methods do not all support metric pose-goal conditioning. Each policy is evaluated over ten zero-shot real-robot trials, with five trials from each of two initial headings (A and B; Fig.~\ref{fig:baselines}).
\coolname{} consistently reaches and stops at the goal in all ten trials, achieving a mean final displacement of $0.31\pm0.11$\,m, compared with OmniVLA's closest approach of $1.45\pm0.38$\,m. GNM generally remains at least $3$\,m away from the goal, while ViNT and NoMaD exhibit greater sensitivity to initial orientation, approaching the goal reliably from only one of the two headings. The same policy also reaches and stops at the image goal in all five trials on the unseen WAVE ROVER, achieving a final displacement of $0.64\pm0.12$\,m. These results demonstrate more consistent image-goal navigation, despite training on a substantially smaller dataset. \coolname{} jointly predicts waypoint distributions and navigation states, enabling goal termination without 2D pose-goal conditioning.

\subsection{Zero-Shot Pose-Conditioned Navigation}
\label{sec:robot_results}

We next extend the evaluation to more challenging navigation settings using 2D pose and image+2D pose goal conditioning. Unlike GNM, ViNT, and NoMaD, OmniVLA supports these additional goal modalities, allowing a direct comparison with \coolname{} as navigation complexity increases. Table~\ref{tab:modality} summarizes the results over five trials per method.
The results reveal a substantial difference when navigation requires deviating from the direct goal direction. In the \emph{Detour Course} ($90^\circ$ start), under image+2D pose conditioning, \coolname{} successfully navigates around the obstacles and reaches the goal in all five trials, achieving a final displacement of $0.40\pm0.03$\,m, compared with OmniVLA's closest approach of $5.13\pm0.10$\,m. While OmniVLA turns toward the goal but fails to navigate around the obstacles, \coolname{} generates intermediate waypoints that enable detouring while maintaining progress toward the goal. A similar difference is observed under
2D pose-only conditioning, with \coolname{} achieving $0.42\pm0.02$\,m compared with $3.23\pm1.98$\,m for OmniVLA across five trials.
On the WAVE ROVER, which has no built-in collision avoidance, \coolname{} also completes all image+pose \emph{Detour Course} trials, achieving a final displacement of $0.43\pm0.03$\,m.

\begin{figure*}[t]
  \centering
  \begin{overpic}[width=\linewidth]{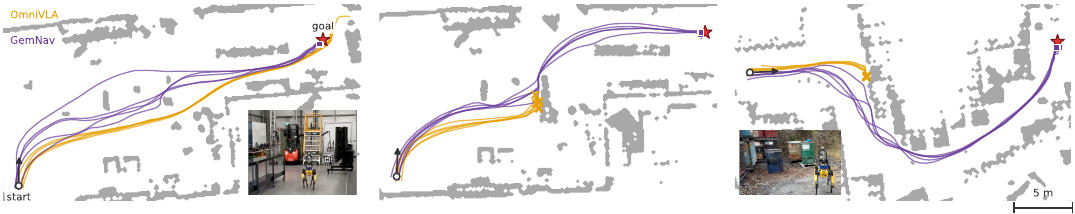}
    \put(16,3){\colorbox{white}{\footnotesize (a)}}
    \put(50,3){\colorbox{white}{\footnotesize (b)}}
    \put(83,3){\colorbox{white}{\footnotesize (c)}}
  \end{overpic}
  \vspace{-16pt}
  \caption{Image+2D pose navigation on Spot (five trials per policy). (a) \emph{Warehouse} (Clear), (b) \emph{Warehouse} (blocked), and (c) \emph{Chemical Yard}. Gray indicates obstacles; squares and crosses denote goal stops and obstacle contacts.} %
  \label{fig:long-courses}
  \vspace{-8pt}
\end{figure*}

This performance gap increases on longer and more challenging navigation routes (Fig.~\ref{fig:long-courses}). In the \emph{Warehouse}, blocking an otherwise clear aisle has little effect on \coolname{}'s FDE ($0.46\pm0.03$\,m versus $0.44\pm0.03$\,m), whereas OmniVLA's closest approach increases from $0.51\pm0.07$\,m to $15.46\pm0.12$\,m. Importantly, this comparison retains the same platform, goal modality, and target, isolating the effect of increased obstacle complexity. In the \emph{Chemical Yard}, \coolname{} reaches and stops at the goal in all five trials, achieving $0.49\pm0.02$\,m compared with OmniVLA of $16.54\pm0.02$\,m.
The results also demonstrate effective navigation without a goal image. Under 2D pose-only conditioning in the \emph{Chemical Yard}, \coolname{} achieves a final displacement of $0.44\pm0.03$\,m across five trials, compared $15.89\pm0.30$\,m in OmniVLA. Together, these findings indicate that the learned metric waypoint policy can maintain goal-directed navigation through successive local detours under both 2D pose and image+2D pose conditioning.

\subsection{Zero-Shot Off-Road Navigation}
\label{sec:robot_offroad}

\begin{figure}[th!]
  \centering
  \includegraphics[width=0.85\linewidth]{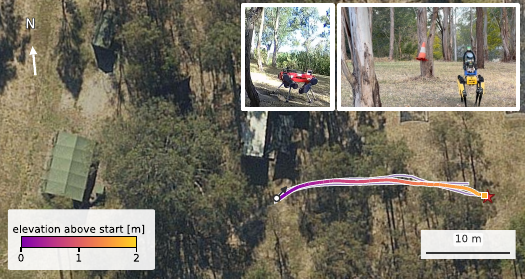}
\caption{Zero-shot off-road navigation on natural terrain using ANYmal and Spot with image+2D pose goals. Goals (stars) and stops (squares) are marked.}
  \label{fig:bushland}
  \label{fig:bushland-map} %
\end{figure}

We further evaluate \coolname{} in unseen natural environments, extending beyond the structured urban and industrial settings considered previously. Using the same SCAND-trained checkpoint, we deploy \coolname{} on ANYmal and Spot over irregular natural terrain with dense vegetation and trees as obstacles. Under image+2D pose conditioning, \coolname{} successfully reaches the goal in all seven uninterrupted ANYmal trials across two goal locations, with final displacements of $ 0.44 \pm 0.07$,m. Similarly, Spot reaches the goal in both uninterrupted trials (Fig.~\ref{fig:bushland}), with a final displacement of approximately $0.43 \pm 0.05$\,m. We additionally evaluate the unseen WAVE ROVER on the same natural terrain using a 2D-pose goal. The rover successfully stops at the goal in two of three trials, with final displacements of $0.33$ and $0.51$\,m, while passing within $0.88$\,m of the goal without stopping in the remaining trial.
These results demonstrate zero-shot transfer to natural environments without fine-tuning. We further evaluate generalization to natural environments on the large-scale WildCross dataset~\cite{wildcross2026} in the following subsection.

\subsection{Offline Evaluation}
\label{sec:offline_results}

Table~\ref{tab:offline-baselines} evaluates offline trajectory prediction by comparing eight predicted local waypoints against their corresponding recorded trajectories across held-out SCAND, TokenWalker, and WildCross datasets. Across all datasets and goal modalities, \coolname{} consistently achieves lower ADE/FDE than the baselines. Under image-goal navigation, it outperforms GNM, ViNT, NoMaD, and OmniVLA, including on WildCross, where it achieves an ADE/FDE of $1.81/3.17$\,m compared with $2.41$--$2.95$\,m ADE and $4.32$--$5.36$\,m FDE for the baselines.

Despite some degradation under visual distribution shift, \coolname{} maintains its performance advantage over the baselines across all evaluation datasets.  Under 2D pose conditioning, ADE remains between $0.37$ and $0.78$\,m, with FDE below $0.20$\,m across all three datasets, substantially outperforming OmniVLA. Combining image and pose goals further reduces ADE across all datasets. These results demonstrate consistent short-horizon trajectory prediction accuracy across goal modalities and unseen environments, complementing the closed-loop real-robot evaluations.

\begin{table}[t]
  \centering
  \footnotesize
  \setlength{\tabcolsep}{3pt}
  \renewcommand{\arraystretch}{0.9}
  \caption{Offline eight-waypoint prediction errors (ADE/FDE in meters) across three datasets and goal modalities. Lower is better.}
  \label{tab:offline-baselines}
  \begin{tabular}{@{}lccc@{}}
    \toprule
    Model & Image & 2D Pose & Image + 2D Pose \\
    \midrule
    \multicolumn{4}{@{}l}{\textit{SCAND (in-distribution)}} \\
    GNM       & $2.83$ / $5.48$ & -- & -- \\
    ViNT      & $2.89$ / $5.56$ & -- & -- \\
    NoMaD     & $2.58$ / $4.61$ & -- & -- \\
    OmniVLA   & $2.55$ / $4.63$ & $2.56$ / $4.70$ & $2.53$ / $4.64$ \\
    \coolname{} & $\mathbf{0.76}$ / $\mathbf{1.20}$ & $\mathbf{0.37}$ / $\mathbf{0.18}$ & $\mathbf{0.33}$ / $\mathbf{0.19}$ \\
    \midrule
    \multicolumn{4}{@{}l}{\textit{TokenWalker}} \\
    GNM       & $2.68$ / $5.22$ & -- & -- \\
    ViNT      & $2.71$ / $5.21$ & -- & -- \\
    NoMaD     & $2.40$ / $4.31$ & -- & -- \\
    OmniVLA   & $2.63$ / $4.77$ & $2.76$ / $4.93$ & $2.54$ / $4.60$ \\
    \coolname{} & $\mathbf{2.09}$ / $\mathbf{3.64}$ & $\mathbf{0.78}$ / $\mathbf{0.19}$ & $\mathbf{0.75}$ / $\mathbf{0.19}$ \\
    \midrule
    \multicolumn{4}{@{}l}{\textit{WildCross (off-road)}} \\
    GNM       & $2.80$ / $5.35$ & -- & -- \\
    ViNT      & $2.78$ / $5.29$ & -- & -- \\
    NoMaD     & $2.41$ / $4.32$ & -- & -- \\
    OmniVLA   & $2.95$ / $5.36$ & $2.76$ / $4.98$ & $2.67$ / $4.81$ \\
    \coolname{} & $\mathbf{1.81}$ / $\mathbf{3.17}$ & $\mathbf{0.50}$ / $\mathbf{0.18}$ & $\mathbf{0.39}$ / $\mathbf{0.18}$ \\
    \bottomrule
  \end{tabular}
\end{table}

 \section{Ablations}
\label{sec:ablations}

\noindent \textbf{Action Representation:}
\label{subsec:abl-head}
We study the contribution of the proposed probabilistic categorical representation against a continuous $L_1$ regression head using the same pretrained Gemma-4 backbone, LoRA configuration, training data, and training protocol. While the categorical head represents each waypoint coordinate as a distribution over discrete metric value bins and soft-decodes it into a continuous coordinate, the $L_1$ head directly regresses continuous waypoint coordinates.

\begin{figure}[t]
  \centering
  \includegraphics[width=\linewidth]{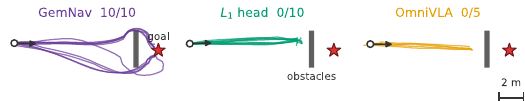}
  \caption{Detour Course, image+2D pose goal, start facing the goal with the obstacle line (gray) across the route. \coolname{} passes the line on either side and stops at the goal; the $L_1$ head and OmniVLA drive straight at the obstacle.}
  \label{fig:obstacle-course}
  \label{fig:obstacles} %
\end{figure}

This distinction matters when multiple navigation trajectories are feasible, such as passing an obstacle on either side. On the \emph{Detour Course}, \coolname{} successfully navigates around the obstacle and stops at the goal in all ten trials, achieving a final displacement of $0.48\pm0.07$\,m. In contrast, the matched $L_1$ regression head drives into the obstacle in all ten trials (Fig.~\ref{fig:obstacle-course}), producing trajectories consistent with collapsing alternative feasible paths into a direct but obstructed route. The OmniVLA, which uses a continuous action head trained with an $L_2$ loss, exhibits similar behavior in all trials, with a closest approach of $3.10\pm0.09$\,m. This shows that the proposed categorical action representation outperforms direct continuous regression for navigation with multiple feasible trajectories.

\noindent \textbf{Effect of Visual History Depth:}
\label{subsec:abl-history}
We investigate the effect of incorporating past visual observations into the visual context, using $H\in\{0,1,2,4\}$ history frames sampled at $1$\,s intervals, where $H=0$ denotes the single-frame policy. All variants follow the same training protocol.
On ANYmal, $H=1$ and $2$ complete all five trials on the \emph{Detour Course}, including a more challenging configuration with an extended obstacle line, where the single-frame policy succeeds in only $3/5$ trials. In contrast, $H=4$ fails to navigate around the obstacles in all five trials and reaches the image-only goal in only $3/10$ trials. Increasing history depth also increases inference latency, from approximately $0.5$\,s for $H=0$ to $1.0$\,s for $H=4$, reducing the frequency of waypoint replanning and navigation-state updates. These results indicate that visual history can improve navigation in challenging settings, whereas longer histories compromise performance.

\section{Conclusion}
\label{sec:conclusion}

This paper introduced \coolname{}, a visual navigation policy that adapts a pretrained MLLM using LoRA and a probabilistic categorical representation of robot-local metric waypoints.
Combined with metric-distance goal sampling, \coolname{} navigates across multiple goal modalities using only 8.7 hours of SCAND demonstrations.
Closed-loop experiments demonstrate zero-shot transfer across three robot platforms and unseen environments, including obstacle courses and natural off-road terrain, and offline evaluations show that \coolname{} outperforms baselines across three datasets and all goal modalities.
Future work will investigate cross-embodiment transfer to additional robot platforms, more efficient integration of visual history, and improved goal termination under image-only conditioning without metric goal coordinates.

\balance{}

  \bibliography{bibfile}  %

  \clearpage

\end{document}